\documentclass[preprint,12pt,nopreprintline]{elsarticle}

\usepackage{amssymb}
\usepackage{lineno}
\usepackage{booktabs}
\usepackage[font=footnotesize,labelfont=bf]{caption}
\usepackage[subrefformat=parens]{subcaption}
\usepackage{array}
\usepackage{bm}
\usepackage{amsmath}

\usepackage{color}

\journal{Pattern Recognition}

\begin{document}

\begin{frontmatter}

\title{TextBoost: Boosting Scene Text Fidelity in Ultra-low Bitrate Image Compression}

\author[aff1]{Bingxin Wang}
\ead{wang.bingxin@huawei.com}
\ead{bwangbo@connect.ust.hk}
\author[aff2]{Yuan Lan}
\ead{lanyuan5@huawei.com}
\author[aff2]{Zhaoyi Sun}
\ead{sun.zhaoyi1@huawei.com}
\author[aff1]{Yang Xiang\corref{cor1}}
\ead{maxiang@ust.hk}
\author[aff2]{Jie Sun}
\ead{j.sun@huawei.com}

\address[aff1]{Department of Mathematics, The Hong Kong University of Science and Technology (HKUST), Clear Water Bay, Sai Kung District, Hong Kong SAR, China}
\address[aff2]{Huawei Hong Kong Research Center, 8/F, 2 Science Park West Avenue, Hong Kong Science Park, Sha Tin, Hong Kong SAR, China}
\cortext[cor1]{Corresponding author}

\begin{abstract}
  Ultra-low bitrate image compression faces a critical challenge: preserving small-font scene text while maintaining overall visual quality. 
  Region-of-interest (ROI) bit allocation can prioritize text but often degrades global fidelity, leading to a trade-off between local accuracy and overall image quality. 
  Instead of relying on ROI coding, we incorporate auxiliary textual information extracted by OCR and transmitted with negligible overhead, enabling the decoder to leverage this semantic guidance.
  Our method, TextBoost, operationalizes this idea through three strategic designs: (i) adaptively filtering OCR outputs and rendering them into a guidance map; (ii) integrating this guidance with decoder features in a calibrated manner via an attention-guided fusion block; and (iii) enforcing guidance-consistent reconstruction in text regions with a regularizing loss that promotes natural blending with the scene. Extensive experiments on TextOCR and ICDAR 2015 demonstrate that TextBoost yields up to 60.6\% higher text-recognition F1 at comparable Peak Signal-to-Noise Ratio (PSNR) and bits per pixel (bpp), producing sharper small-font text while preserving global image quality and effectively decoupling text enhancement from global rate-distortion optimization.
\end{abstract}

\end{frontmatter}


\section{Introduction}
\label{submission}
Image compression is fundamental to reducing the storage and transmission burden of images. 
Recent advances in deep learning have enhanced both rate-distortion efficiency and perceptual quality, surpassing traditional methods \citep{wang2003multiscale,balle2016end,balle2018variational,mentzer2020high,dubois2021lossy,he2022elic,xu2022multi, gao2022flexible, liu2023learned,lieberman2023neural, guo2023compression, ali2023towards,lee2024neural,xu2024idempotence,jia2024generative}. 
In bandwidth-constrained scenarios such as satellite communication, images must be compressed to ultra-low bitrates while still maintaining sufficient visual fidelity. 
Achieving such extreme compression ratios without perceptual degradation remains a major challenge, especially for preserving fine-grained details.
In recent years, substantial progress has been made in learned image compression for ultra-low bitrate applications. Several works~\cite{pan2022extreme, lei2023textsketchimage, careil2023towards, yang2023lossy, relic2024lossy} have leveraged diffusion models to improve perceptual quality. 
At ultra-low bitrates, these methods adopt diffusion models as decoders, leveraging their strong generative capacity, together with conditional information (e.g., text captions), to generate reconstructed images that are perceptually close to the original.

Despite these advances, preserving the fidelity of small-font text remains an under-explored challenge in ultra-low bitrate scenarios. This issue is especially critical in mission-critical applications (e.g., search and rescue or surveillance) where such text often conveys essential information. A conventional approach is to allocate more bits to text regions through ROI coding~\cite{uchigasaki2023deep}, but this strategy increases transmission cost and often degrades global perceptual quality. This inherent trade-off between localized accuracy and global visual quality poses a structural limitation for existing approaches.

We take a different perspective. Modern OCR systems reliably extract textual content and spatial position information, which can be compressed and transmitted with negligible overhead. 
Transmitting OCR text strings requires far fewer bits than compressing equivalent image content.
This auxiliary information can thus be used to enhance small-font text in reconstructed images while avoiding the drawbacks of ROI methods.

However, effectively integrating this auxiliary information remains non-trivial. 
A straightforward idea is to overlay the OCR text onto the reconstructed image according to the recognized coordinates.
However, under extreme compression, text regions in the decoded image often appear blurry, distorted, or inconsistent with surrounding structures, which means that directly inserting OCR outputs cannot recover the fine-grained appearance, style, and alignment needed for text to blend naturally into the scene.
Our key insight is that auxiliary textual information should act as guidance rather than a replacement for visual reconstruction, preserving both textual fidelity and visual coherence.

Building on this insight, we articulate three design principles for text-aware compression at ultra-low bitrates: (1) treat OCR as a lightweight semantic prior rather than a pixel source; (2) render recognized strings and boxes into a geometrically aligned guidance map; and (3) couple this guidance to the reconstruction stream in a calibrated manner. The guidance map must be visually coherent and robust to recognition noise, orientation, and scale. When applied to the reconstruction stream appropriately, it should preserve global appearance while letting text blend naturally into the scene.

To this end, we introduce TextBoost, a framework specifically designed to operationalize these principles. First, we adaptively filter and compress only the OCR content and geometry that contribute most to small-text fidelity. Second, we decode this information into a guidance map that is robust to rotation and recognition noise. Third, we integrate the guidance into the decoder through an attention-guided fusion block and a guidance-consistent loss. This mechanism balances the auxiliary signals with learned representations, ensuring that text regions are sharpened while preserving global image quality, without requiring explicit bitrate reallocation.

Extensive experiments on the TextOCR dataset \citep{singh2021textocr} demonstrate that TextBoost improves the F1 score by up to 60.6\% over state-of-the-art methods at comparable PSNR and bpp. This gain in text recognizability confirms the effectiveness of our strategy in decoupling text enhancement from global rate-distortion optimization. 
When OCR information is unavailable or unreliable, the framework gracefully degrades to the underlying image codec without introducing artifacts or additional failure modes.

\begin{figure}
\centering
\includegraphics[width=\linewidth]{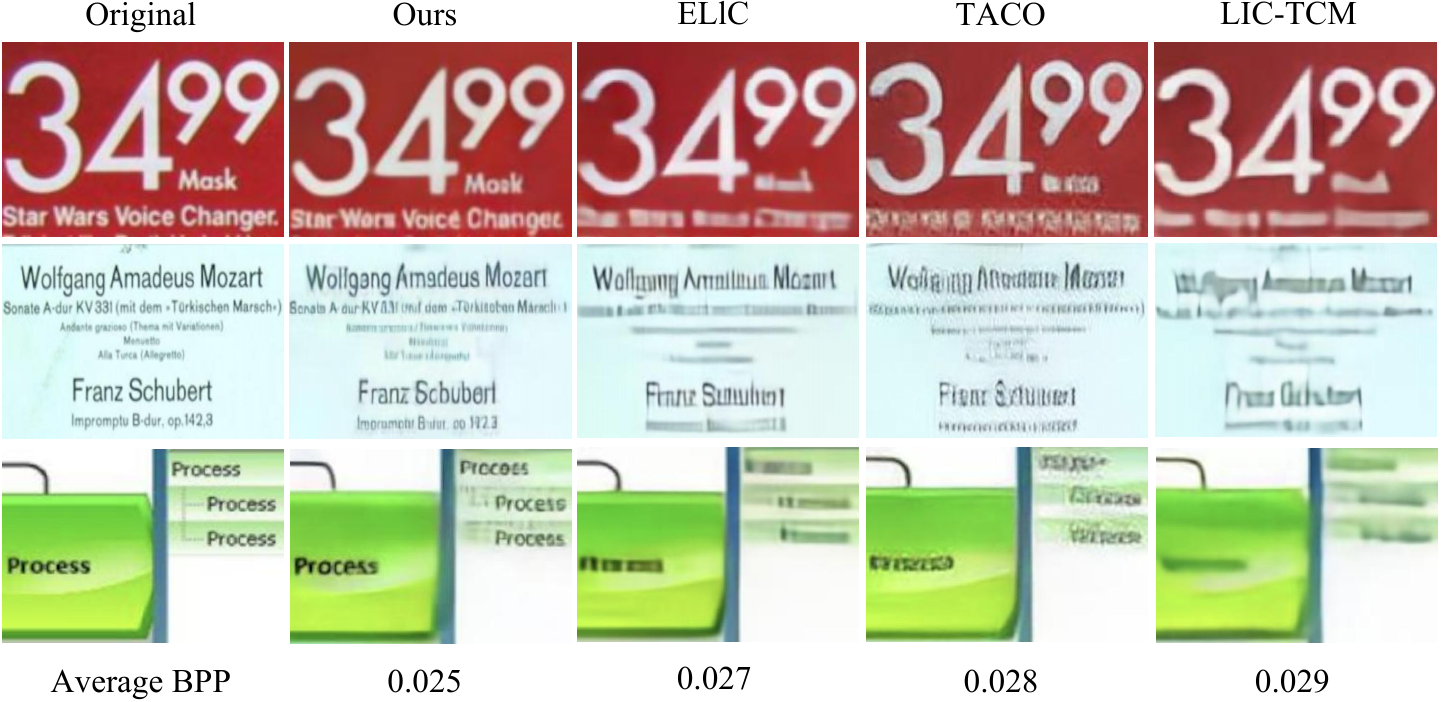}
\caption{\textbf{TextBoost preserves small-text fidelity at ultra-low bitrates.} 
Visual comparison on the TextOCR validation set \citep{singh2021textocr} against 
ELIC \citep{he2022elic}, TACO \citep{lee2024neural}, and LIC\mbox{-}TCM \citep{liu2023learned}. 
Our method delivers clearly better text reconstruction at similar or lower bitrates, with improved preservation of fine typographic details.}
\label{performance}
\label{fig:intro}
\end{figure}

\section{Related Works}

\subsection{Learned Image Compression}

End-to-end neural codecs have become increasingly prominent in lossy compression, starting with the integration of quantization techniques and autoencoder architectures \citep{balle2016end}. Ballé et al.'s pioneering work \citep{balle2018variational} introduces a hyperprior to capture latent spatial dependencies effectively. The authors of \cite{minnen2018joint} combine autoregressive and hierarchical priors to exploit latent probabilistic structures. Further advancements include compression with discretized Gaussian mixture likelihoods and attention modules \citep{cheng2020learned}, as well as compression with unevenly grouped space-channel contextual adaptive coding \citep{he2022elic}. Other research \citep{agustsson2019generative, mentzer2020high} incorporates generative models into learned compression for more realistic reconstructions. LIC-TCM \citep{liu2023learned} introduces an efficient Transformer-CNN Mixture (TCM) block and a channel-wise entropy model with Swin-transformer-based attention. TACO \citep{lee2024neural} presents a text-guided compression method, leveraging the ELIC \citep{he2022elic} architecture to achieve both high pixel-level and perceptual quality without relying on generative models.
However, these methods are not specifically optimized for preserving small-font text, producing blurry text regions as shown in Fig.~\ref{performance}  and Fig.~\ref{performance2}.

\subsection{Low Bitrate Learned Image Compression}

Recent advancements in learned image compression at low bitrates have explored various techniques. Beyond the aforementioned methods \citep{Zhou_2018_CVPR_Workshops,muckley2023improving}, generative models have gained significant attention. In \cite{pan2022extreme}, the authors incorporate text embeddings as conditional inputs to a diffusion-based decoder to enhance perceptual quality, while \cite{lei2023textsketchimage}'s authors leverage ControlNet \citep{zhang2023adding} to condition their decoder on both text and sketches, improving semantic and perceptual performance. However, these approaches face inherent challenges due to the rate-distortion-realism tradeoff \citep{blau2018perception}. The loss of pixel fidelity can limit their practicality, and the stochastic nature of generative models compromises fine-grained detail reconstruction.

Given these limitations, our study specifically addresses detail-sensitive applications (e.g., small-text preservation). Rather than employing generative frameworks, we push the boundaries of conventional LIC to maximize textual fidelity while maintaining compression efficiency.

\subsection{ROI-Based Text Enhancement in Image Compression}

ROI techniques are a long-standing direction in learned image compression. Classical ROI-based codecs allocate higher quality to selected regions using binary masks \citep{agustsson2019generative, cai2019end}, and later variants employ spatially varying quality maps for finer control \citep{song2021variable}.

This idea has recently been applied to scene text. Uchigasaki et al.~\cite{uchigasaki2023deep} enhance text regions by modulating local quality levels through a learned quality map, improving readability under constrained bitrates. However, ROI-based approaches remain fundamentally limited by the trade-off between local fidelity and global perceptual quality.
Unlike these methods that simply redistribute the bit budget, TextBoost introduces explicit semantic guidance from an auxiliary stream. This strategy effectively decouples text enhancement from rate allocation, allowing for improved recognizability without sacrificing the quality of non-text regions.

\section{Background}
This section provides an overview of the key concepts: learned lossy image compression and the text spotting model for evaluating reconstructed images.
\subsection{Learned Lossy Image Compression}
\label{3.1}
Learned lossy image compression \citep{balle2016end,balle2018variational,minnen2018joint} optimizes the rate–distortion tradeoff using deep neural networks. The process typically involves encoding an input image \( \boldsymbol{x} \) into a compact latent representation, which is then quantized, entropy‑coded, transmitted, entropy‑decoded, and finally reconstructed. Specifically, the image \( \boldsymbol{x} \) is first encoded by the encoder \( g_a \) into a latent representation:
\begin{equation}
\boldsymbol{y} = g_a(\boldsymbol{x})
\end{equation}
The latent feature \( \boldsymbol{y} \) is then quantized into \( \hat{\boldsymbol{y}} \) to efficiently store or transmit the compressed representation. To improve the encoding process, a hyperprior \( \hat{\boldsymbol{z}} \) is introduced \citep{balle2018variational}, which captures spatial dependencies in the latent representation by a parametric entropy model. First, \( \boldsymbol{y} \) is fed into the hyperprior encoder \( h_a \) to generate \( \boldsymbol{z} \); then, \( \boldsymbol{z} \) is quantized to obtain \( \hat{\boldsymbol{z}} \). The hyper decoder $ h_s $ uses $ \hat{\boldsymbol{z}} $ to estimate the distribution parameters, which are utilized for the entropy encoding and decoding of $ \hat{\boldsymbol{y}} $, thereby providing side information.

This also allows for a more accurate estimation of the conditional probability distribution of \( \hat{\boldsymbol{y}} \), modeled as a conditioned Gaussian:
\begin{equation}
p_{\hat{\boldsymbol{y}} \mid \hat{\boldsymbol{z}}}(\hat{\boldsymbol{y}} \mid \hat{\boldsymbol{z}}) = \mathcal{N}(\boldsymbol{\mu}, \sigma^2) * U(-0.5, 0.5),
\end{equation}
where \( \mathcal{N}(\boldsymbol{\mu}, \sigma^2) \) represents a Gaussian with mean \( \boldsymbol{\mu} \) and variance \( \sigma^2 \). The variance \( \sigma^2 \) is derived from the hyperprior \( \hat{\boldsymbol{z}} \). \( U(-0.5, 0.5) \) introduces small uniform noise to enhance robustness. The expected negative entropy of this probability mass function \( p_{\hat{\boldsymbol{y}}} \), \( -\mathbb{E}[\log p_{\hat{\boldsymbol{y}}}(\hat{\boldsymbol{y}})] \), serves as the rate term.

\( \hat{\boldsymbol{y}} \) is then passed to the decoder \( g_s \), which reconstructs the image \( \boldsymbol{\hat{x}} \):
\begin{equation}
\boldsymbol{\hat{x}} = g_s(\hat{\boldsymbol{y}})
\end{equation}
The full process—encoding, quantization, entropy coding, transmission, entropy decoding, and reconstruction—enables efficient image compression. The training objective is a rate–distortion loss that balances the bitrate \( R(\hat{\boldsymbol{y}}) \) and the reconstruction error \( D(\boldsymbol{x}, \boldsymbol{\hat{x}}) \):
\begin{equation}
\mathcal{L} = R(\hat{\boldsymbol{y}}) + \lambda D(\boldsymbol{x}, \boldsymbol{\hat{x}})
\end{equation}
The estimated bitrate \( R(\hat{\boldsymbol{y}}) \) is computed using the entropy model, where the negative log‑probability of \( \hat{\boldsymbol{y}} \) is used to calculate the rate:
\begin{equation}
\label{losseq}
R(\hat{\boldsymbol{y}}) = -\mathbb{E}[\log p_{\hat{\boldsymbol{y}}}(\hat{\boldsymbol{y}})]
\end{equation}

\subsection{End-to-End Text Spotting Model}
End-to-end scene text spotting refers to the simultaneous detection and recognition of text in images within a unified framework. Unlike traditional OCR pipelines that separate text detection and recognition into independent stages, end-to-end models jointly optimize both tasks, enabling better contextual understanding and higher efficiency. These models leverage deep neural networks, often incorporating CNN-based or Transformer-based architectures, to directly extract textual content from images.  

A key challenge in scene text spotting is handling diverse text shapes, orientations, and layouts, especially in complex real-world environments. Recent advancements have introduced models that generalize across multiple detection formats. One such approach is the UNIfied scene Text Spotter (UNITS) \citep{kil2023towards}, which unifies the detection of both quadrilateral and polygonal text regions, allowing it to recognize text in arbitrary shapes. Additionally, UNITS employs a starting-point prompting mechanism, enabling the model to detect a number of text instances exceeding the maximum quantity encountered during training, thus improving robustness in real-world scenarios.  

In this paper, we evaluate the recognizability of text in reconstructed images by extracting OCR information using a pre-trained text spotting model. The extracted text is then compared with ground truth annotations to compute the F1 score, which directly reflects the usability of reconstructed text in downstream applications. We use the F1 score at matched bitrates to compare the performance of different compression methods in small-font text reconstruction.

\begin{figure}[t!]
\centering
\includegraphics[width=\linewidth]{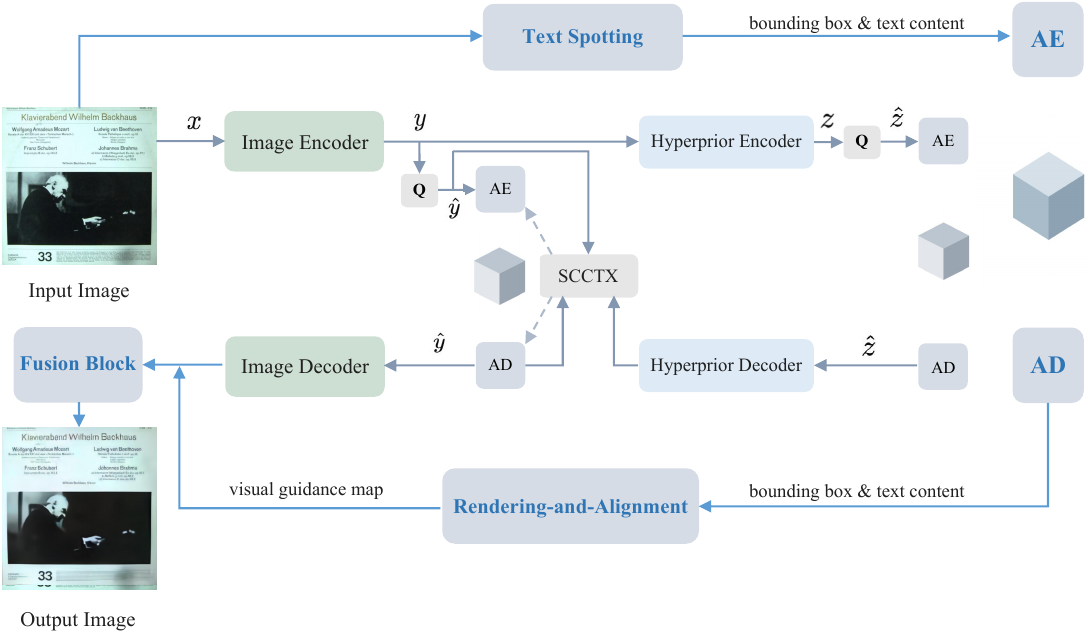}
\caption{\textbf{Overall pipeline of TextBoost.} The framework is built upon a learned image compression backbone (comprising the Image Encoder, Hyperprior network, and Image Decoder). In parallel, the text branch extracts and transmits OCR information, which is processed by the Rendering-and-Alignment module to generate a visual guidance map. Finally, the Fusion Block integrates this guidance with the features decoded by the baseline network to produce the final output. SCCTX refers to the space-channel contextual model.}
\label{fig24}
\end{figure}

\section{Method}

Motivated by the limitations of ROI and the potential of OCR guidance outlined in Section~\ref{submission}, we now address the practical challenge of designing a system that effectively leverages this auxiliary information.
The core difficulty lies in bridging the semantic gap between discrete textual content and continuous visual representations while maintaining both textual accuracy and visual coherence under extreme compression constraints.

TextBoost operationalizes the design principles introduced in Section~\ref{submission} through a holistic pipeline comprising three key modules: (1) an adaptive rendering-and-alignment module that converts OCR content and geometry into a geometrically aligned guidance map (Section~\ref{4.1decode}); (2) a feature fusion module that integrates auxiliary text representations with decoder outputs (Section~\ref{4.2fusion}); and (3) a guidance-consistent loss that promotes faithful blending of text regions into the reconstructed scene (Section~\ref{4.3loss}). The complete pipeline is illustrated in Fig.~\ref{fig24}.

\subsection{Adaptive OCR Information Processing: From Text to Visual Guidance}
\label{4.1decode}
We transmit only the most informative textual information and convert it into visual guidance for the reconstruction pathway. We treat OCR as a lightweight semantic prior: rather than pixels, we send content and geometry, then render this information into a geometry- and scale-aligned guidance map. Transmitting all detected text is inefficient, as large-font text is generally robust to compression artifacts. Moreover, even with perfect textual content and positions, we still need to transform discrete strings into visual representations that can guide reconstruction.

Our approach begins with adaptive filtering. Given OCR information including text box coordinates $\{b_1, \dots, b_n\}$ and their corresponding content $\{c_1, \dots, c_n\}$, we calculate the average character area $\overline{A}_i$ for each text instance and selectively transmit only those with $\overline{A}_i$ below a threshold. This prioritization is motivated by a key observation: small-font text suffers most from compression artifacts, while large-font text remains relatively legible even at low bitrates.

The more challenging task is converting the filtered text information into visual guidance that can assist reconstruction. Simply overlaying text onto a black canvas would create harsh boundaries and unrealistic appearances that could mislead the fusion process.

\begin{figure}[!t]
\centering
\includegraphics[width=\linewidth]{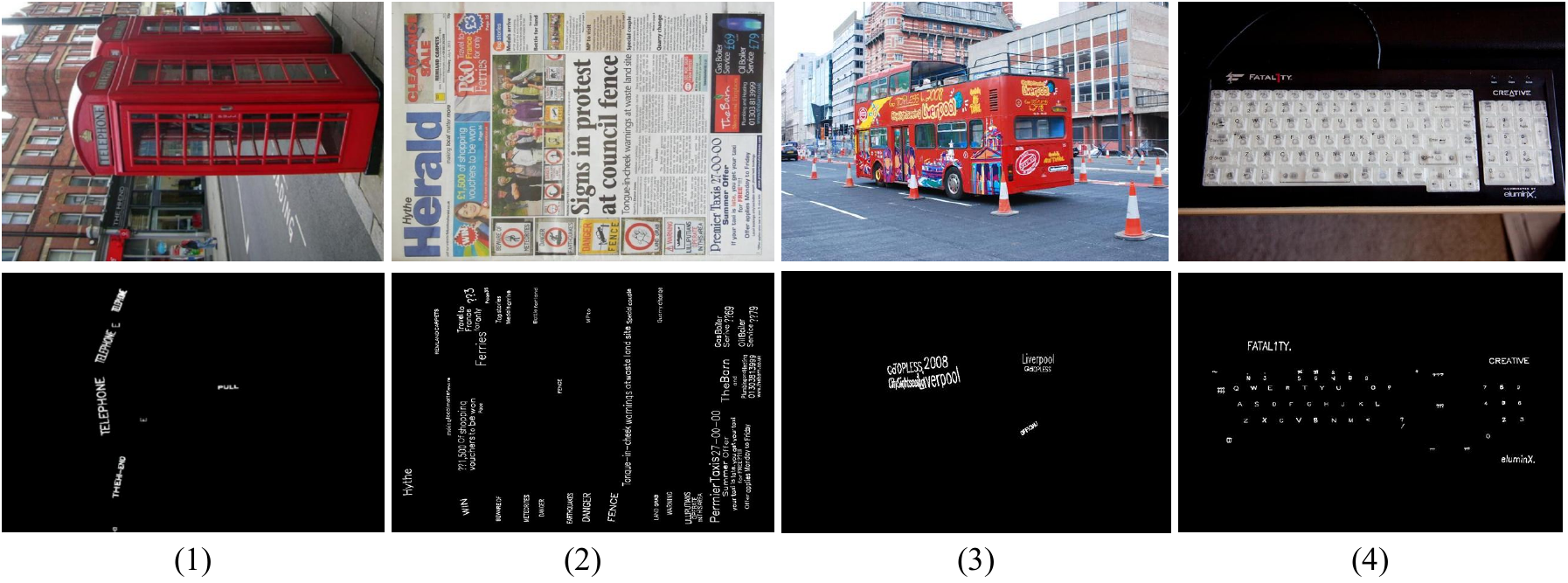}
\caption{\textbf{Examples of guidance maps generated from OCR information on the TextOCR dataset \citep{singh2021textocr}.} The first row shows the original scene images and the second row the corresponding rendering-and-alignment results. (1) and (3) illustrate robust handling of diverse in-plane orientations. By internally normalizing text to a horizontal layout, our method accurately renders text at distinct angles (e.g., vertical in (1) and slanted in (3)). (2) and (4) demonstrate selective transmission and precise rendering: large-font text is filtered out according to the average character-area criterion, while small-font content is rendered with accurate geometry and spatial placement.}
\label{bwi}
\end{figure}

To address this, we introduce an adaptive text rendering algorithm that creates visually coherent auxiliary images. The filtered OCR data $\{b_i, c_i\}$ is compressed using a standard lossless text compression scheme (gzip in our implementation) and transmitted to the decoder, where our algorithm reconstructs it into visual guidance through several carefully designed steps. 

First, we compute the minimal bounding rectangle for each text box and estimate the in-plane orientation to normalize the text to a left-to-right horizontal layout. Concretely, given a quadrilateral or polygonal box $b_i=\{(x_k, y_k)\}$, we determine the rotation angle $\theta$ as the angle between the horizontal axis and the principal edge direction (i.e., the longer side of the box or a least-squares-fitted baseline through the vertices). We then rotate the crop by $-\theta$ to obtain a horizontally aligned reference before rendering. Next, we adaptively adjust font sizes to fit text within the designated regions, ensuring that the auxiliary image captures the spatial extent of text areas. The text is rendered on a black background, forming clear spatial masks that indicate where text should appear.

Importantly, when no OCR information is available, we output a zero tensor, ensuring that our method gracefully degrades to standard compression without introducing artifacts. After rendering, we rotate the text back to its original orientation and crop to match target dimensions, creating an auxiliary image that spatially aligns with the original while providing clear textual guidance.

The effectiveness of this adaptive rendering process is demonstrated in Fig.~\ref{bwi}, where we show how diverse scene text is converted into clean auxiliary representations that preserve spatial relationships while providing clear guidance for the fusion module.

\begin{figure}[!t]
\centering
\includegraphics[width=\linewidth]{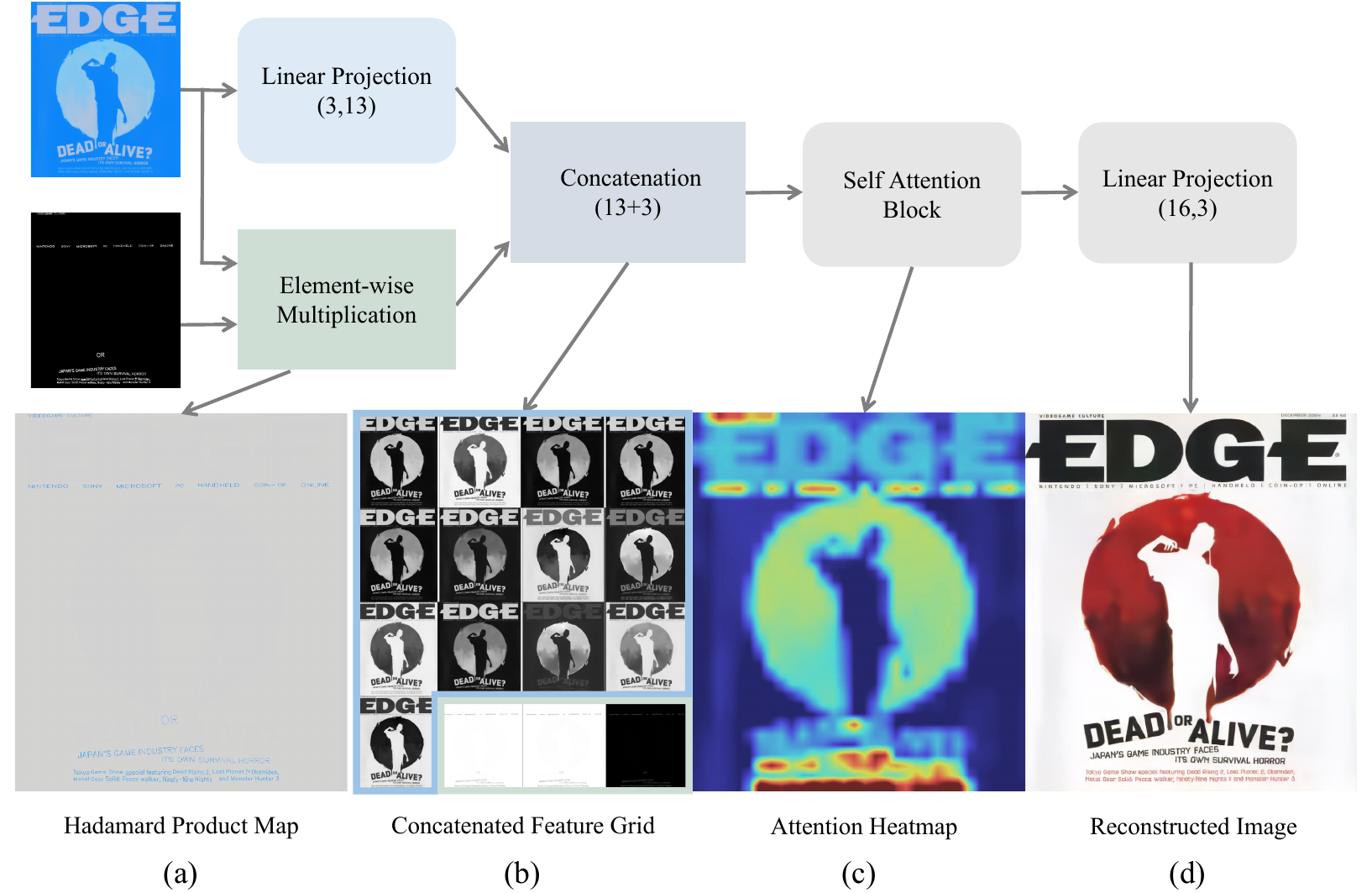}
\caption{\textbf{Fusion block architecture and intermediate visualizations.} Top: the proposed fusion block couples the auxiliary guidance with decoder features through element-wise modulation, channel expansion and concatenation, an attention module, and a final 1×1 projection to RGB. Bottom: visualizations aligned with each stage. (a) \emph{Hadamard product map}: element-wise multiplication where white glyphs in the auxiliary map inherit color from the decoder features. (b) \emph{Concatenated features (13 + 3)}: the first 13 channels (blue branch) are projected from the decoder output, while the last 3 channels (green branch) correspond to the Hadamard product map. (c) \emph{Attention heatmap}: activations of the attention block, mostly concentrated on small-font text areas. (d) \emph{Reconstruction}: the final image after the 1×1 projection.}
\label{fig:fusion_block}
\end{figure}

\subsection{Adaptive Feature Fusion: Balancing Guidance and Fidelity}
\label{4.2fusion}

We integrate the auxiliary guidance into the reconstruction stream in a calibrated manner: the fusion block \emph{modulates} the decoder features rather than replacing them, so non‑text appearance continues to be governed by the learned image prior.

Our fusion strategy in TextBoost, illustrated in Fig.~\ref{fig:fusion_block}, is driven by the principle that auxiliary text information should guide the reconstruction while remaining faithful to the scene statistics. Since the auxiliary image provides reliable spatial layouts but lacks fine textures,  naively incorporating it risks introducing artifacts. We therefore couple it to the decoder features through a carefully designed attention mechanism.

Specifically, the fusion process begins by embedding the guidance into the decoder feature space. We compute a Hadamard product between the auxiliary map and the decoder RGB output, ensuring that glyph pixels inherit color information from the decoder features. To accommodate this guidance while preserving representational capacity, we expand the decoder output from 3 to 13 channels through a 1×1 convolution and concatenate it with the 3-channel modulated guidance map. This results in a comprehensive 16-channel representation (13 + 3), as depicted in Fig.~\ref{fig:fusion_block}(b).

Subsequently, these concatenated features are processed by an attention module adapted from \citep{cheng2020learned}. This module, comprising stacked 1×1 and 3×3 convolutions, learns spatial‑channel weights to explicitly emphasize small‑font regions while suppressing irrelevant responses. This step ensures global coherence by integrating broader contextual information. Finally, a 1×1 convolution projects the refined features back to the 3‑channel RGB space, producing a reconstructed image that exhibits both enhanced text fidelity and a natural appearance.

\subsection{Guidance-Consistent Loss: Preserving Text Fidelity}
\label{4.3loss}
To ensure the network enhances text rather than simply copying auxiliary patterns, we regularize training with a guidance-consistent objective. Without appropriate constraints, the fusion module may take shortcuts by directly imprinting auxiliary text onto the image, which leads to artifacts and poor integration with surrounding visual content.

In practice, we observe a divergence: while the semantic recognizability of small-font text improves, the pixel-level fidelity (PSNR) in these regions may drop below the global average. To prevent this imbalance, we introduce a guidance-consistent loss that encourages the fused reconstruction to maintain text-region fidelity on par with the global level—without altering rate allocation or adding inference-time overhead.

Let $m$ be a binary mask built from OCR-derived annotation boxes on the training set, and define the guidance-consistent loss $\mathcal{L}_{\text{gc}}$ as:
\begin{equation}
  \mathcal{L}_{\text{gc}} = \mathrm{MSE}(m \odot x,\ m \odot \hat{x}).
  \label{eq:lgc_def}
\end{equation}
Crucially, this design decouples text enhancement from rate control. Unlike Region-of-Interest (ROI) methods that redistribute bits, we freeze all upstream modules (encoder, hyperprior, and the decoder base) during the optimization of $\mathcal{L}_{\text{gc}}$. This confines the gradient updates solely to the fusion block, ensuring that improvements stem from better feature integration rather than increased bit consumption in text regions.

To formally implement this decoupling, we adopt a two-stage training strategy:

\textit{Stage 1: Standard Rate-Distortion Optimization.} Initially, the entire framework is trained using the conventional rate–distortion objective to optimize the fundamental compression performance:
\begin{equation}
  \mathcal{L}_{\text{stage1}} = \mathcal{L}_{\text{bpp}} + \lambda \cdot \mathrm{MSE}(x,\hat{x}).
\end{equation}

\textit{Stage 2: Text-Aware Fusion Fine-tuning.} We then freeze the encoder, the entropy model, and the decoder base (layers preceding the fusion block). The optimization is confined strictly to the fusion block, minimizing a joint objective that promotes text consistency without altering the rate distribution:
\begin{equation}
  \mathcal{L}_{\text{stage2}} = \overline{\mathcal{L}_{\text{bpp}}} + \lambda (\mathrm{MSE}( x,\hat{x}) + \alpha \, \mathcal{L}_{\text{gc}}).
\label{eq:stage2_loss}
\end{equation}
Here, $\overline{\mathcal{L}_{\text{bpp}}}$ indicates that the bitrate term remains constant during this phase.
Note that while $\lambda$ does not affect the optimal solution, we retain it to maintain consistent gradient magnitudes compatible with the Stage 1 learning dynamics.
The coefficient $\alpha$ is calibrated so that text-region fidelity is commensurate with the image-wide level.

\section{Experiments}
\subsection{Experimental Setup}

\subsubsection{Datasets}
We evaluate our method on three standard datasets covering both text-centric and general image compression scenarios:

(1) \textbf{TextOCR} \citep{singh2021textocr}. We use the TextOCR training set for fine-tuning, which comprises 21,778 images with 714,770 word annotations (including bounding boxes and transcripts). For evaluation, we utilize the validation split containing 3,124 images. This dataset provides diverse, dense, and challenging text regions, making it ideal for assessing text preservation.

(2) \textbf{ICDAR 2015} \citep{karatzas2015icdar}. The test set consists of 500 images capturing multi-oriented text in complex real-world scenes. The variety of fonts, sizes, and layouts allows us to test the robustness and generalization of our method on unseen data.

(3) \textbf{Kodak} \citep{kodak}. This dataset includes 24 natural images commonly used for benchmarking general image compression. Unlike the text-focused datasets above, Kodak serves to verify that our method maintains high-quality reconstruction on general scenes without text annotations. Results on Kodak are presented in Table~\ref{kodak_table}.

\subsubsection{Baseline Models}
We compare TextBoost against two classical standards—JPEG and VTM \citep{bross2021overview}—and four state-of-the-art learned methods: ELIC \citep{he2022elic}, LIC-TCM \citep{liu2023learned}, TACO \citep{lee2024neural}, and MS-ILLM \citep{muckley2023improving}. 
ELIC and LIC-TCM serve as strong pixel-fidelity baselines. Since TextBoost is built upon the ELIC backbone, direct comparison highlights the gain from our feature fusion. TACO represents methods using region-based guidance, while MS-ILLM employs generative techniques for low-bitrate consistency.
For fair comparison, we fine-tune the models of ELIC, LIC-TCM, and TACO on our TextOCR training set to target the same ultra-low bitrate range. 
For MS-ILLM, we follow the official evaluation protocol and directly evaluate the released pre-trained checkpoints. 
While the authors provide training code, dataset-specific fine-tuning is not part of the standard evaluation pipeline for the GAN-based MS-ILLM models, which are primarily released for direct inference at target bitrates.

\subsubsection{Measurements}
The total bitrate is calculated as the sum of the compressed image stream and the auxiliary stream (comprising encoded OCR coordinates and text strings) to ensure a fair comparison.
For image quality, we report PSNR (pixel-wise fidelity), LPIPS \citep{zhang2018unreasonable} (perceptual quality), and MS-SSIM \citep{wang2003multiscale} (structural similarity).
To assess machine-readability, we run the high-accuracy text spotting model $\mathrm{UNITS}_{\text {Shared }}$ \citep{kil2023towards} on the reconstructed images. We report the F1 scores for both text detection (DET) and end-to-end recognition (E2E), which provide a comprehensive measure of text recoverability.

\begin{figure}
\centering
\includegraphics[width=\linewidth]{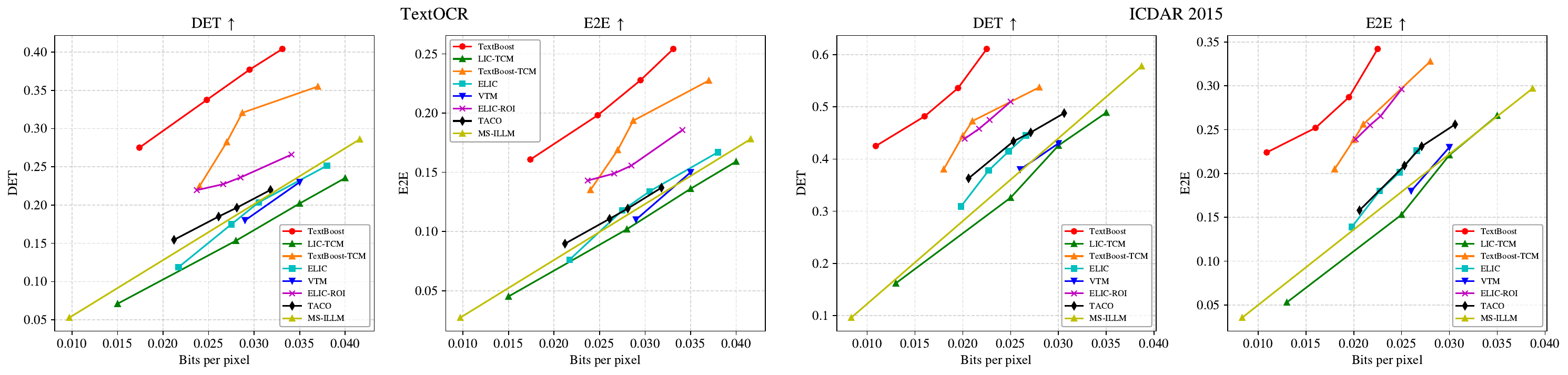}
\caption{\textbf{Quantitative evaluation of text spotting performance.} Rate-distortion curves showing Text Detection (DET) and End-to-End Recognition (E2E) F-measures on TextOCR \citep{singh2021textocr} and ICDAR 2015 \citep{karatzas2015icdar}. TextBoost (red curves) yields significant improvements over state-of-the-art learned compression methods (e.g., LIC-TCM, ELIC) and ROI-based baselines, demonstrating superior text preservation capabilities at ultra-low bitrates.}
\label{fig5}
\end{figure}

\subsubsection{Implementation Details}
Following the two-stage protocol described in Sec.~\ref{4.3loss}, we first pre-train the backbone compression framework on ImageNet \citep{deng2009imagenet} for 1000 epochs using the standard rate-distortion loss. This phase utilizes the Adam optimizer with a learning rate of $10^{-4}$ and a batch size of 32. 
During data loading, we randomly crop training images to $256 \times 256$. Text annotations with an average character area exceeding $T_{\text{train}} = 150$ are filtered out to focus on small-text reconstruction.

In the second stage, we fine-tune on the TextOCR training set for 20 epochs using the joint objective defined in Eq.~(\ref{eq:stage2_loss}). As designed, we freeze the encoder, entropy model, and decoder base, updating only the fusion-related modules under the guidance-consistent constraint. These trainable components are optimized with a higher learning rate of $10^{-3}$ to facilitate rapid convergence.

Throughout the experiments, the guidance-consistent loss weight is set to $\alpha = 10$. For evaluation on full-resolution scene images, we adjust the filtering threshold to $T_{\text{test}} = 800$ to account for the resolution disparity compared to training crops. The fine-tuning process requires approximately 24 hours on hardware with an aggregate computational throughput of around 160 TFLOPs (FP16 precision).

\begin{figure}
\centering
\includegraphics[width=\linewidth]{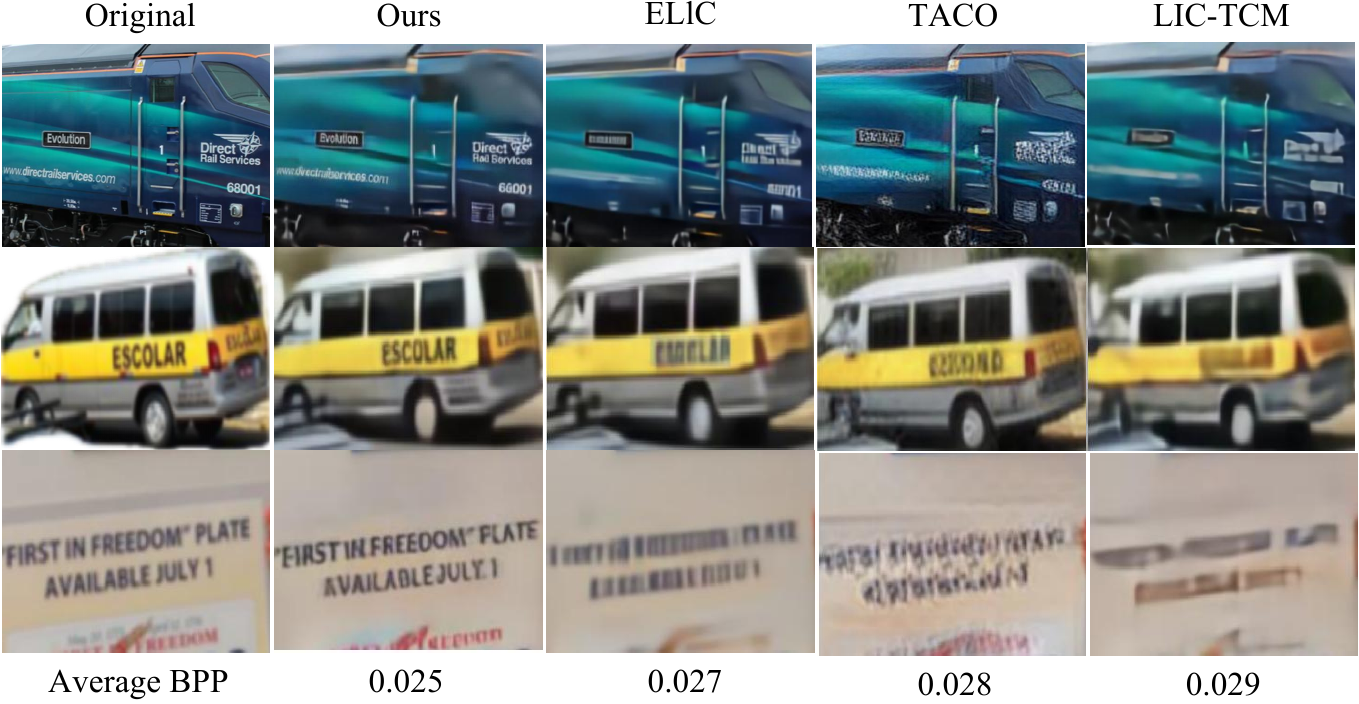}
\caption{\textbf{Additional visual comparisons across diverse street scenes on TextOCR.} TextBoost produces sharper glyphs and cleaner fine strokes while retaining the global scene appearance.Notably, our method achieves this superior text fidelity at a lower average bitrate (0.025 bpp) compared to the baselines (0.027–0.029 bpp).}
\label{performance2}
\end{figure}

\subsection{Results}

\subsubsection{Quantitative Performance on Text Spotting}
As illustrated in Fig.~\ref{fig5}, TextBoost consistently outperforms all baselines across both the TextOCR and ICDAR 2015 datasets. 
On TextOCR, our method achieves state-of-the-art performance in both detection and recognition tasks. Specifically, at similar or lower bitrates ($\approx 0.033$ bpp), TextBoost achieves a DET F1 score of 0.404, representing a \textbf{60.6\% relative improvement} over the best baseline, ELIC (0.2515). Notably, our method also significantly surpasses the ROI-based variant (ELIC-ROI), confirming that explicit semantic guidance is more effective than simple bit reallocation.
This advantage generalizes well to the ICDAR 2015 dataset. At 0.0225 bpp, TextBoost achieves an E2E score of 0.342, outperforming ELIC (0.180) by 90\%. These results suggest that TextBoost effectively decouples text recognizability from the rate-distortion trade-off.

\subsubsection{Global Quality Consistency}
\begin{figure}[!t]
\centering
\includegraphics[width=0.95\linewidth]{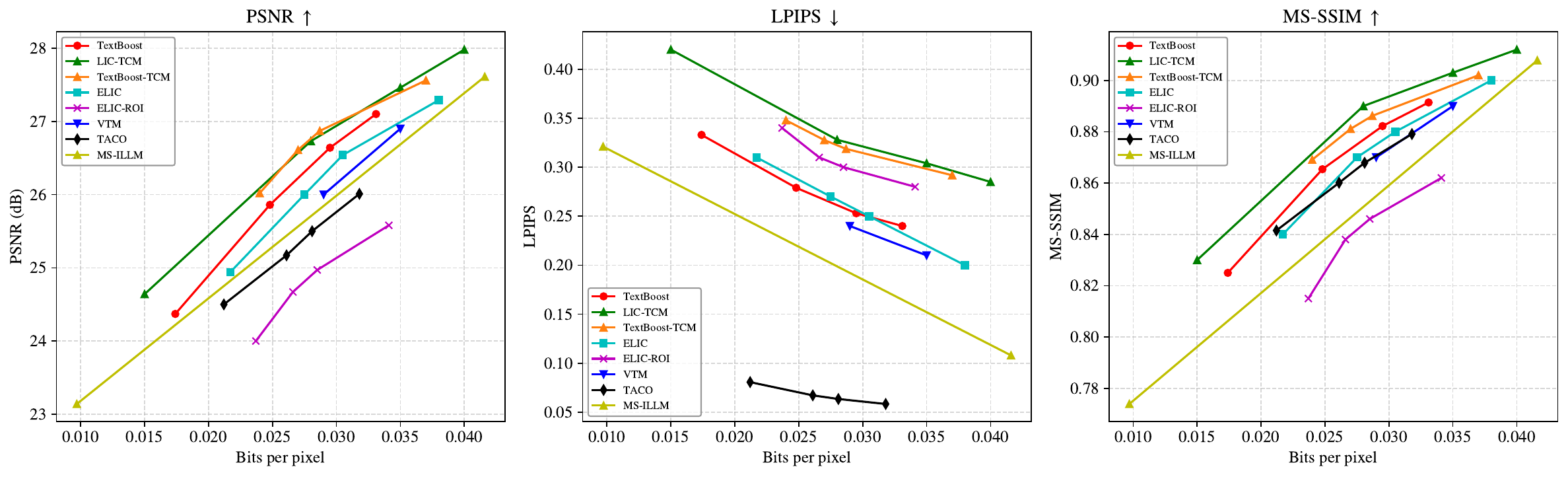}
\caption{\textbf{Global quality consistency on TextOCR: PSNR, LPIPS, and MS-SSIM.} Comparison of rate-distortion curves against state-of-the-art baselines. TextBoost maintains competitive global fidelity, achieving PSNR/MS-SSIM comparable to ELIC and LIC-TCM while offering better perceptual quality (lower LPIPS). With the LIC-TCM backbone (\emph{TextBoost-TCM}), our method effectively matches the original LIC-TCM in reconstruction metrics, demonstrating that our text enhancement incurs negligible cost to global scene statistics.}
\label{textocr_overall}
\end{figure}

\begin{figure}[!t]
\centering
\includegraphics[width=0.95\linewidth]{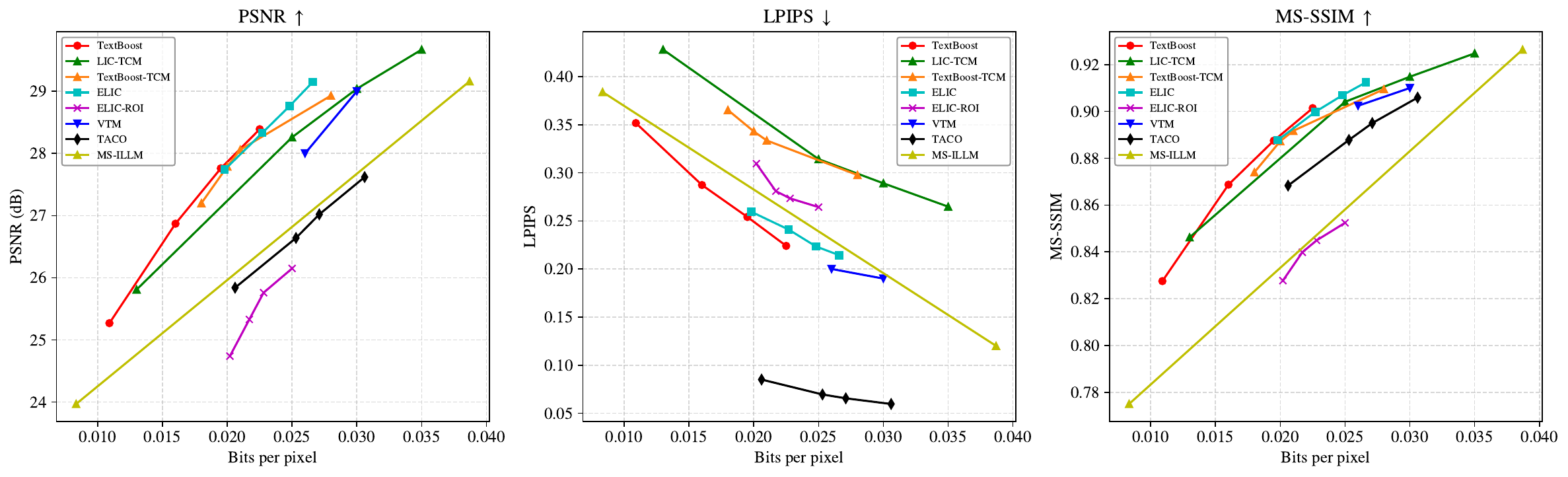}
\caption{\textbf{Generalization to ICDAR 2015: PSNR, LPIPS, and MS-SSIM.} Consistent with TextOCR results, TextBoost maintains high global quality on the ICDAR 2015 dataset \citep{karatzas2015icdar}. Notably, TextBoost achieves strong PSNR and LPIPS performance at ultra-low bitrates, indicating that the proposed guidance-based fusion mechanism generalizes well across different scene distributions.}
\label{ic_overall}
\end{figure}

Beyond text improvements, TextBoost maintains competitive global image fidelity.
On TextOCR (Fig.~\ref{textocr_overall}), our method achieves PSNR and MS-SSIM scores comparable to the state-of-the-art LIC-TCM and ELIC, while delivering improved perceptual quality (lower LPIPS).
Furthermore, on the ICDAR 2015 dataset (Fig.~\ref{ic_overall}), TextBoost demonstrates strong robustness, achieving leading PSNR and highly competitive MS-SSIM at ultra-low bitrates compared to existing baselines. This indicates that our feature fusion mechanism not only preserves text but also positively influences the reconstruction of complex scene structures without introducing artifacts.

\subsubsection{Visual Comparisons}
Fig.~\ref{performance2} presents visual examples from the validation set. TextBoost produces visibly clearer small-font text with sharper strokes and intact typographic details compared to baselines. Importantly, as indicated by the metrics in Fig.~\ref{fig5}, our method achieves this superior text fidelity at a lower average bitrate (0.025 bpp) than competing methods (0.027--0.029 bpp), demonstrating its high coding efficiency.

\subsubsection{Backbone Transferability}
To verify the portability of our approach, we apply the TextBoost strategy to the LIC-TCM codec, denoted as \emph{TextBoost-TCM}. As shown in the Rate-Distortion curves, TextBoost-TCM closely tracks the original LIC-TCM in terms of PSNR and MS-SSIM but yields a substantial gain in OCR F1 scores. This confirms that our decoder-side guidance is model-agnostic and can be seamlessly integrated into different architectures to boost small-text recognizability.

\subsubsection{Comparison with ROI Baseline}
We also evaluate an ROI-based scheme (ELIC-ROI) that allocates more bits to text regions. While this scheme improves text-specific scores compared to vanilla ELIC, it noticeably degrades the overall image quality (as seen in the LPIPS curves in Fig.~\ref{textocr_overall}), creating a direct trade-off between text and background. In contrast, TextBoost improves text fidelity without sacrificing global appearance.

\subsubsection{Generalization to Non-Text-Centric Images}
To validate robustness on general content, we evaluate on the Kodak dataset (Table~\ref{kodak_table}). TextBoost achieves the highest PSNR (24.70 dB) among all learned baselines at similar bitrates, outperforming both ELIC and LIC-TCM. While LIC-TCM yields a marginally higher MS-SSIM, our method remains highly competitive and demonstrates better perceptual quality than LIC-TCM (LPIPS: 0.42 vs. 0.55). Crucially, these results confirm that incorporating text-aware enhancements does not degrade general image reconstruction capabilities on natural scenes lacking dense text.

\begin{table}[h]
\centering
\caption{\textbf{Performance comparison on the Kodak dataset.} Our method achieves the best reconstruction quality in terms of PSNR and maintains competitive structural (MS-SSIM) and perceptual (LPIPS) fidelity compared to state-of-the-art baselines.}
\label{kodak_table}
\begin{tabular}{lcccc}
\toprule
\textbf{Method} & \textbf{bpp} & \textbf{PSNR} & \textbf{MS-SSIM} & \textbf{LPIPS} \\
\midrule
ELIC   & 0.027 & 24.22 & 0.8110 & 0.42 \\
TACO   & 0.028 & 23.04 & 0.7720 & \textbf{0.09} \\
LIC-TCM   & 0.028 & 24.53 & \textbf{0.8197} & 0.55 \\
TextBoost   & 0.028 & \textbf{24.70} & 0.8150 & 0.42 \\
\bottomrule
\end{tabular}
\end{table}

\subsection{Ablation Study}
\label{ablation_sec}

To validate the effectiveness of the proposed framework, we conduct an ablation study on the TextOCR validation set. We focus on dissecting the contributions of two core components: the Attention-guided Fusion Block and the Guidance-consistent Loss.

\begin{figure}
\centering
\includegraphics[width=0.9\linewidth]{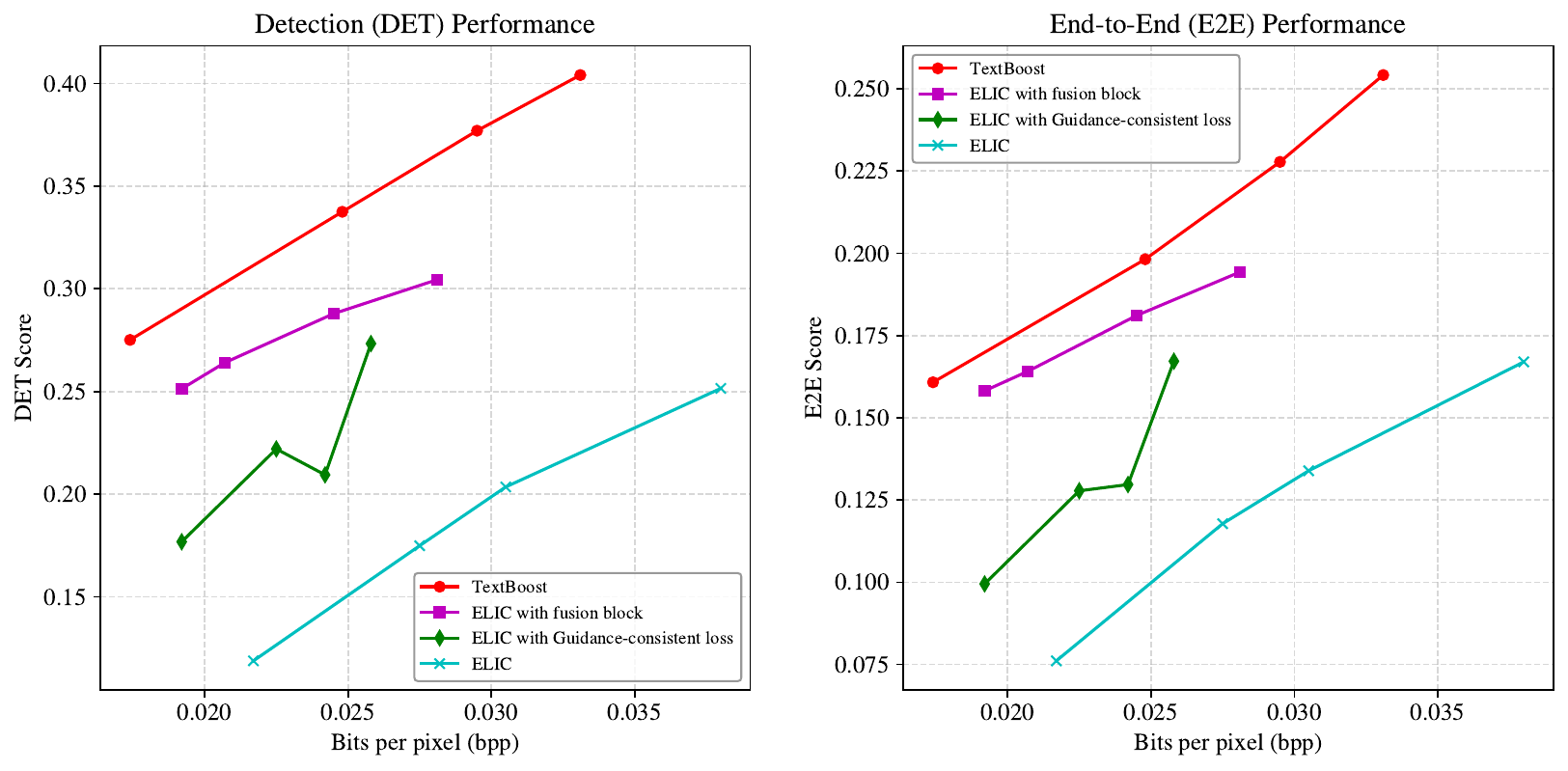}
\caption{\textbf{Ablation study: Effectiveness of key components.} Quantitative comparison of DET and E2E performance on TextOCR. We observe that both the \textbf{Fusion Block} (purple) and the \textbf{Guidance-consistent Loss} (green) individually improve upon the ELIC baseline (cyan). However, the full TextBoost framework (red) yields the highest scores, demonstrating the \textbf{necessity and complementary nature} of combining adaptive feature integration with loss-based constraints.}
\label{fig:ablation}
\end{figure}

\textit{Impact of the Fusion Block:} The fusion block is the primary driver of performance improvement. As illustrated in Fig.~\ref{fig:ablation} (left), simply equipping the baseline with our fusion module (purple curve) yields a substantial performance leap over the ELIC baseline (cyan). For instance, at approximately 0.025 bpp, the fusion-only variant nearly doubles the DET F1 score compared to ELIC (improving from $\sim0.15$ to $\sim0.29$). This confirms that the structural guidance integrated by the fusion module is critical for recovering fine-grained text features, even before specific loss constraints are applied.

\textit{Impact of the Guidance-consistent Loss:} While the fusion block provides the structural foundation, the guidance-consistent loss offers critical regularization. Comparing the full TextBoost (red) with the ``Fusion only'' variant (purple), we observe that excluding this loss leads to a noticeable degradation. Specifically, at 0.025 bpp, the DET and E2E scores decrease by more than 22\% and 18\%, respectively. This demonstrates that explicitly enforcing consistency between the auxiliary guidance and the reconstructed text is essential for maximizing recognition accuracy.

\textit{Conclusion:} These results indicate that the two components are complementary. The fusion block provides the architectural capacity necessary to utilize the OCR priors, while the loss function ensures that these priors are optimally aligned with the reconstruction objectives. Together, they enable TextBoost to achieve the best rate-accuracy trade-off.

\section{Conclusion}
This work offers a new perspective on text preservation in ultra-low bitrate image compression by moving beyond the conventional rate-allocation paradigm. Instead of competing for limited bits between text and non-text regions, we establish that leveraging auxiliary textual information as a lightweight semantic prior effectively guides small-font reconstruction while maintaining global image quality.

Our key insight is that the challenge lies not in \emph{what} to transmit, but in \emph{how} to seamlessly integrate this auxiliary guidance. TextBoost addresses this through a holistic pipeline: adaptive OCR processing that converts discrete strings into visually coherent guidance, and an attention-guided fusion mechanism that modulates decoder features to balance structural fidelity with learned scene statistics.

Experimental evaluations confirm that our approach successfully decouples text enhancement from the traditional rate-distortion trade-off. By achieving substantial gains in text recognizability (up to 60.6\% F1 improvement) with minimal transmission cost, TextBoost paves the way for next-generation content-aware compression systems.

While our current approach demonstrates robust performance on scene text, extending this OCR-guided paradigm to handwritten document image enhancement \citep{jemni2022enhance} presents unique theoretical challenges. Standard OCR extracts semantic content but typically discards the unique stylistic nuances of handwriting, making our current rendering-based guidance strategy insufficient for preserving calligraphic fidelity. Addressing this limitation to capture distinct handwriting styles remains a specific focus for our future research. Beyond the textual domain, the core principle of using auxiliary semantic information as reconstruction guidance extends to other visual elements. Future work could explore similar strategies for critical categories such as faces or objects, or investigate how multiple types of auxiliary information can be jointly leveraged.

\section*{Acknowledgments}
Y. X. was supported in part by the Project of Hetao Shenzhen-HKUST Innovation Cooperation Zone, China HZQBKCZYB2020083 and the Project P2070 at HKUST Shenzhen Research Institute, China.

 \bibliographystyle{elsarticle-num} 
 \bibliography{reference}

\end{document}